\documentclass{article}
\usepackage{spconf,amsmath,epsfig}
\usepackage{times}
\usepackage{epsfig}
\usepackage{graphicx}
\usepackage{amsmath}
\usepackage{amssymb}
\usepackage{amsfonts}
\usepackage{mathrsfs}
\usepackage{bm}
\usepackage{float}
\usepackage{multirow}
\usepackage{color}
\usepackage{url}

\let\OLDthebibliography\thebibliography
\renewcommand\thebibliography[1]{
  \OLDthebibliography{#1}
  \setlength{\parskip}{0pt}
  \setlength{\itemsep}{0pt plus 0.3ex}
}

\begin{document}\sloppy

\def\x{{\mathbf x}}
\def\L{{\cal L}}

\title{HAZY RE-ID: AN INTERFERENCE SUPPRESSION MODEL FOR DOMAIN ADAPTATION PERSON RE-IDENTIFICATION UNDER INCLEMENT WEATHER CONDITION}
%
\name{
Jian~Pang$^{\scriptscriptstyle 1,2, \dagger}$ , Dacheng~Zhang$^{\scriptscriptstyle 1, \dagger}$,
Huafeng~Li$^{\scriptscriptstyle 1,*}$,
Weifeng~Liu$^{\scriptscriptstyle 2}$,
Zhengtao~Yu$^{\scriptscriptstyle 1}$
\thanks{$\dagger$ Equal contribution; * Corresponding author.
}
}

\address{$^{\scriptscriptstyle 1}$Kunming University of Science and Technology;
$^{\scriptscriptstyle 2}$China University of Petroleum (East China)}

\maketitle

\begin{abstract}
In a conventional domain adaptation person Re-identification (Re-ID) task, both the training and test images in target domain are collected under the sunny weather. However, in reality, the pedestrians to be retrieved may be obtained under severe weather conditions such as hazy, dusty and snowing, etc. This paper proposes a novel Interference Suppression Model (ISM) to deal with the interference caused by the hazy weather in domain adaptation person Re-ID. A teacher-student model is used in the ISM to distill the interference information at the feature level by reducing the discrepancy between the clear and the hazy intrinsic similarity matrix. Furthermore, in the distribution level, the extra discriminator is introduced to assist the student model make the interference feature distribution more clear. The experimental results show that the proposed method achieves the superior performance on two synthetic datasets than the state-of-the-art methods. The related code will be released online \url{https://github.com/pangjian123/ISM-ReID}.
\end{abstract}

\section{Introduction}

 With a given image of a target person, the person Re-ID aims at identifying the same person in an image gallery collected from non-overlapping camera views. Due to its importance in video surveillance and security, it arouses widespread research interests in recent years. Although supervised person Re-ID methods have achieved satisfactory results~\cite{Luo2020,xu2020}, they often face significant performance degradation while deploying the trained model to new domains. Domain adaptation person Re-ID requires no labels and can well generalizes on a target dataset. However, the target data used in the training and test of the existing cross-domain methods are mostly collected in sunny weather~\cite{Yang2020,Song2020b,Ge2020,Zhong2019,Cheng2020,Huang2020,xiang2020}, where in practice, the target data used for test may be collected under hazy weather.

\begin{figure}[!t]
	\centering
	\includegraphics[width=3.3in,height=3.3in]{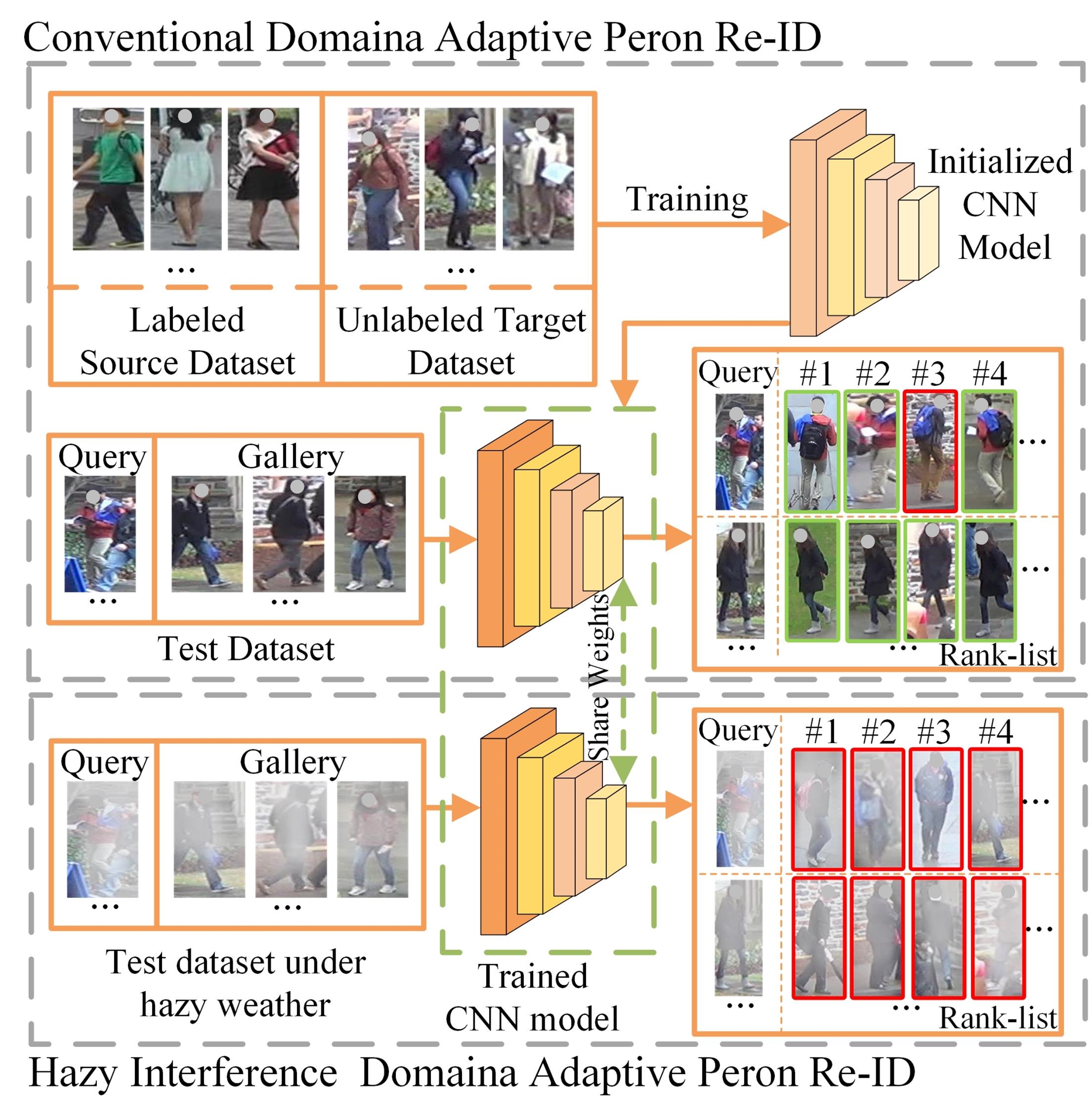}
	\caption{In the conventional domain adaptation person Re-ID task, the data in target domain are collected under sunny weather. If the test data of target domain encounters hazy weather, it will interfere the model to identity pedestrians thus reducing the performance.}
	\label{fig:1}
\end{figure}

 The model trained by the clear images of target domain will cause a significant decline in performance while directly testing on hazy images due to the interference (Fig.\ref{fig:1}). The clear training image and the hazy test image will produce a new domain gap. It is thus difficult to identify the person from the images collected in a hazy weather based only on the common feature representation. Interference always exist in the pedestrian dataset and affect the identification performance of the model. For example, the resolution of the query image and the gallery image do not match~\cite{Cheng2020}, the training image and the test image can belong to different modalities~\cite{Wu2020}, and the pedestrians in the test images can be overlapped~\cite{Zhao2020}. However, those above methods are all trained in a supervised way.

 Domain adaptation person Re-ID has been studied extensively in recent years. These methods can be categorized into two kinds: 1) methods that use Transfer Learning (TL) for domain adaptation~\cite{xiang2020,Zhong2019}, and 2) the ones that use Clustering for Pseudo-label Prediction (CPP)~\cite{Yang2020,Zhang2019,Song2020b,Ge2020}. TL methods usually transfer the image style from the labeled source domain to the target domain. However, the performance highly depends on the quality of the transferred images. CPP predict pseudo-labels for the target samples and select those with high confidence to tune the model by supervised training. Nevertheless, the data used in the mentioned methods are all collected in sunny weather. Most existing methods focus on non-natural disturbances. The scalability of the model will be reduced if the impact of the inclement weather on Re-ID is neglected.

 According to the authors' knowledge, no approach for domain adaptation person Re-ID under hazy weather has been proposed yet. To this end, an Interference Suppression Model (ISM) composed of a teacher-student model and a distribution discriminator is proposed in this work. Under the guidance of the teacher network, the student network can learn robust feature representations that are not affected by hazy interference. ISM constrains the student network to extract features that are not affected by interference information through exploring the inherent feature correlation between pure images and hazy images. The hazy image will impact the original data distribution and the performance. In order to distill the interference information in the data distribution, with the aid of the discriminator, adversarial learning is used to make the student model to extract more pure features with Hazy-Market1501 and Hazy-DukeMTMC-reID datasets. The contributions of this paper are summarized as follows:
\begin{itemize}

\item To the best of our knowledge, we are the first to address the interference problem for domain adaptation person Re-ID under hazy circumstance.

\item We propose a novel model to reduce the interference of hazy on person Re-ID task. Specifically, a teacher-student model is used to learn clear representation under the guidance of the intrinsic similarity matrix. Furthermore, the extra discriminator is involved to make the interference feature distribution closer to the clear feature distribution.

\item Since the previous work did not consider the impact of hazy on domain adaptation person Re-ID. We contribute two synthetic hazy datasets to reduce the negative impact of hazy interference on performance when deployed the trained model to target domain. The experimental results demonstrate that our proposed approach is effective and superior to other methods.

\end{itemize}

\section{Proposed Method}

\begin{figure}[!t]
 	\centering
 	\includegraphics[width=3.3in,height=2.5in]{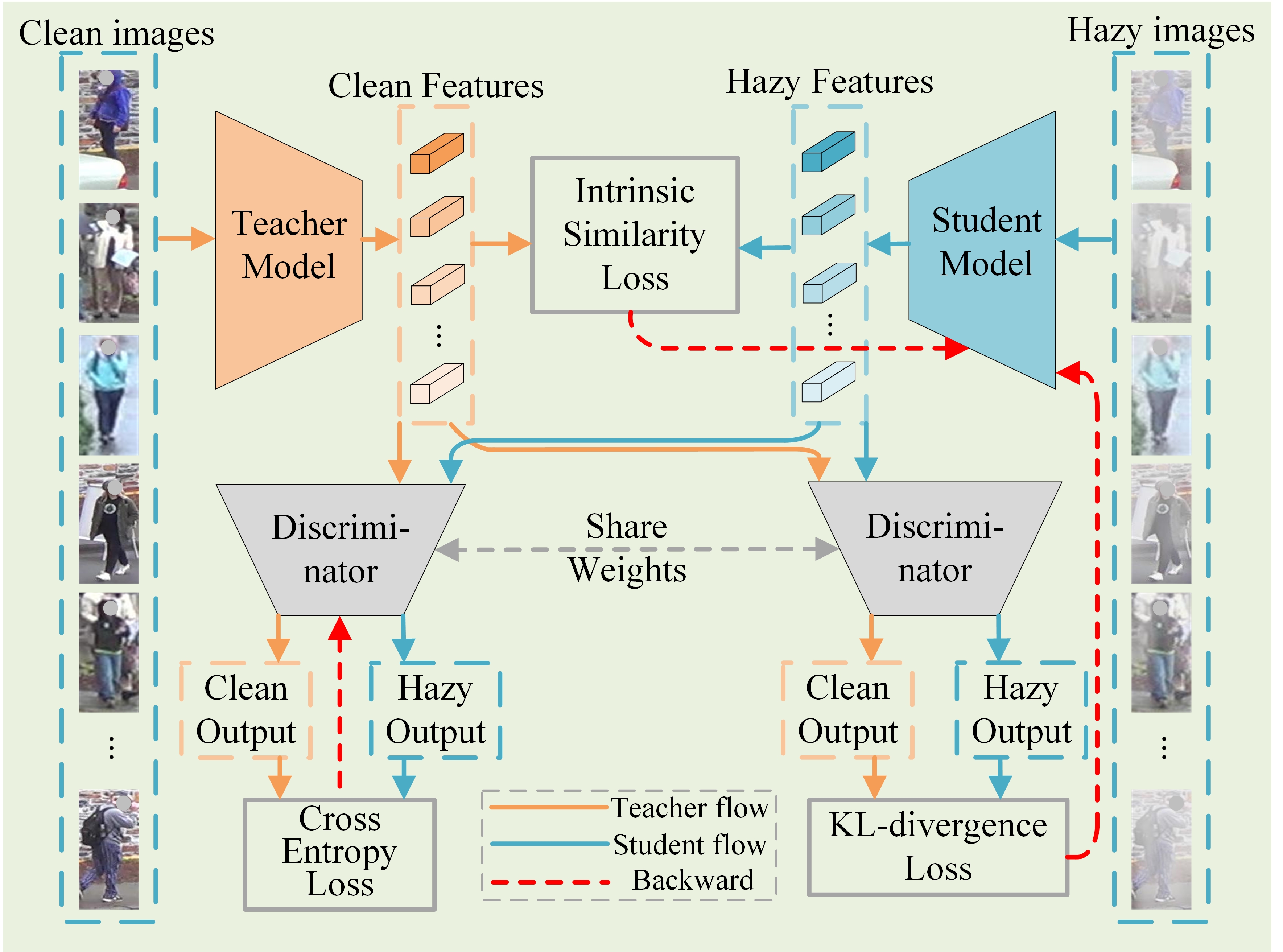}
 	\caption{Framework of the proposed ISM. It includes a teacher-student model and a discriminator. The teacher-student model is pre-trained on source dataset. The intrinsic similarity and interfere distillation by adversarial learning are used to train the student model by fixing the trained teacher model.}
 	\label{fig:2}
 \end{figure}

As shown in Fig.\ref{fig:2}, a novel ISM is proposed to reduce the disturbance of hazy weather on domain adaptation person Re-ID. The hazy weather has a significant impact on identifying the pedestrian whereas being largely ignored by existing methods. The proposed solution is to use the discrepancy of intrinsic similarity matrix between the clean and hazy images to learn a robust feature representation. In addition, the discriminator is involved such that the student model can be guided to distill the interference by aligning distribution through the adversarial training.

\subsection{Supervised Pre-training on Source Domain}
 The labeled source domain dataset $\bm D_{s}=\{\bm X_{s}, \bm Y_{s}\}$ contains $N_{s}$ images of $C$ persons. $\bm X_{s}$ and $\bm Y_{s}$ are the images and person identities, respectively. Each image $x_{i}^{s}$ in $\bm X_{s}$ is associated with an identity label $y_{i}^{s}$ in $\bm Y_{s}$. The target dataset $\bm D_{t}=\{x_{i}^{t}|_{i=1}^{N_{t}}\}$ has no identity label. In order to adapt the model to a hazy weather, we add hazy condition to the training data of the target domain as $\bm D_{h}=\{x_{i}^{h}|_{i=1}^{N_{t}}\}$. $\bm F_{T}$ is the teacher model, $\bm F_{S}$ is the student model, and $\bm D$ is the discriminator. In order to realize a robust feature representation for target domain, the teacher-student model is pre-trained on a labeled source dataset by cross-entropy loss, which can be defined as:
\begin{equation}
\begin{aligned}
    \bm L_{CE}(\bm F_{*})
    =\sum_{i=1}^{C}-(y_{i}^{s}\log\hat{y}_{i}^{s}+(1-y_{i}^{s})\log\hat{y}_{i}^{s})
\end{aligned},
\end{equation}
where $\bm F_{*}$ denotes $\bm F_{T}$ or $\bm F_{S}$, $\hat{y}_{i}^{s}$ represents the predicted logits of class $i$. Refer to~\cite{He2020}, the label smoothing is used to prevent over-fitting. Thus, the $y_{i}^{s}$ in Equation~(1) can be defined as $y_{i}^{s}(i=c)=1-\epsilon$ and $y_{i}^{s}(i\neq c)=\frac{\epsilon}{C-1}$, where c is the ID label corresponding to the input image.

 \begin{figure}[htbp]
 	\centering
 	\includegraphics[width=2.6in,height=2.4in]{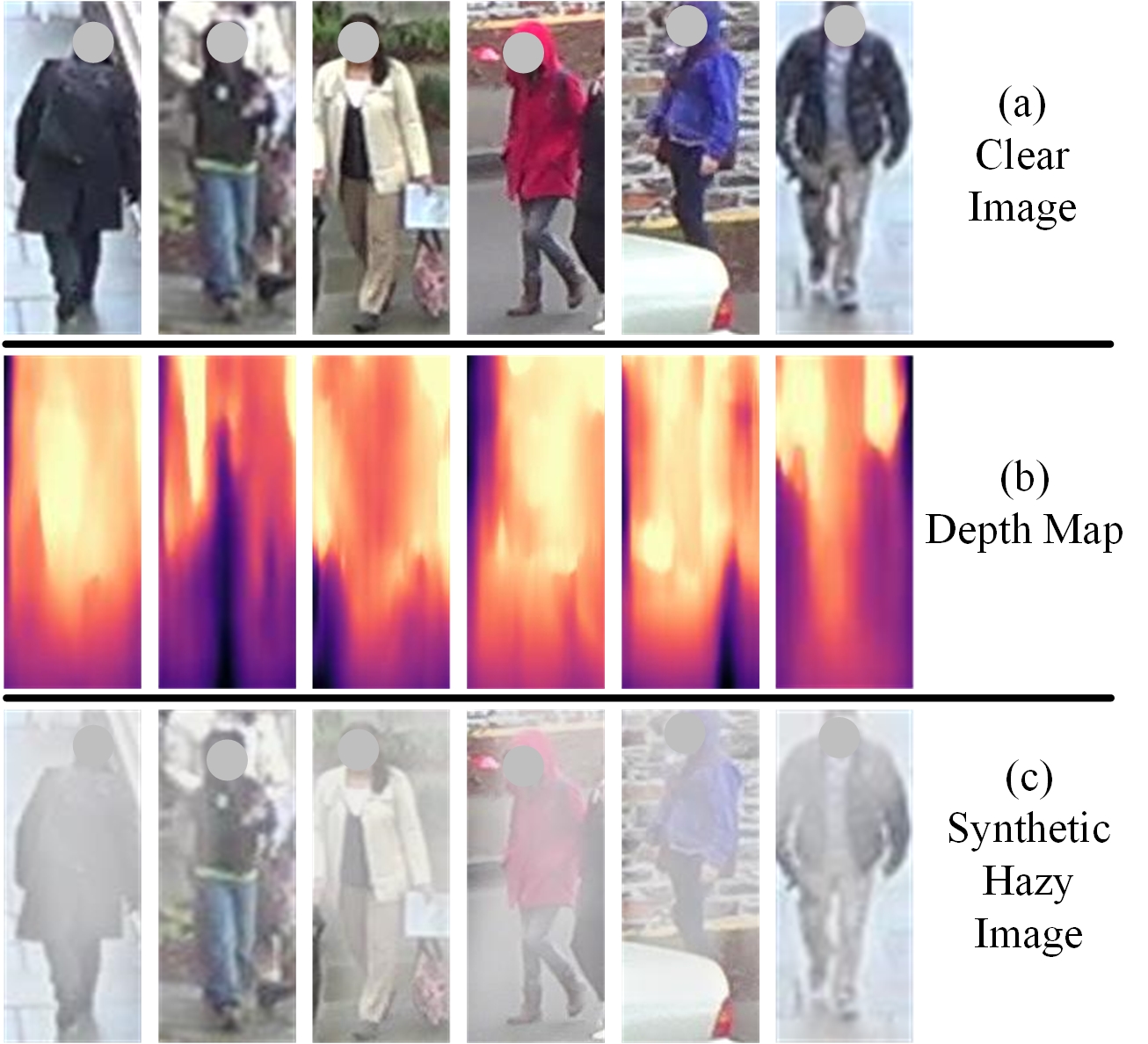}
 	\caption{Visualization of clear images, generated depth maps and synthetic hazy images.}
 	\label{fig:3}
 \end{figure}

\subsection{Intrinsic Similarity Learning}
 After pre-training on the labeled source domain, the teacher model and the student model can have basic recognition capabilities, but lack the ability to adapt to target domain and hazy weather. The proposed ISM aims to learn more robust feature representations that are not affected by hazy-interference for target domain. For the same person image, the feature $\bm f_{i}^{S}=\bm F_{S}(x_{i}^{h})$ generated under the interference will be significantly different from the feature $\bm f_{i}^{T}=\bm F_{T}(x_{i}^{t})$ generated under sunny weather. Based on this assumption, we feed the clear images into the teacher model to calculate the similarity between each other through the distance measurement to obtain an intrinsic similarity matrix, which represents the inner connection between pure images. It can be expressed as:
\begin{equation}
\begin{aligned}
    \bm M_{ij}^{c}=\| \bm f_{i}^{T} - \bm f_{j}^{T}\|_{2},\quad i,j=1,2,\ldots,B
\end{aligned},
\end{equation}
 For the hazy intrinsic similarity matrix, it can be calculated by:
\begin{equation}
\begin{aligned}
    \bm M_{ij}^{h}=\| \bm f_{i}^{S} - \bm f_{j}^{S}\|_{2},\quad i,j=1,2,\ldots,B
\end{aligned},
\end{equation}
 where $B$ is the batch-size. The interference of the hazy will destroy the original connection of the pure image, making the result of the inherent similarity matrix and the pure image have a huge discrepancy. We use the discrepancy between the $\bm M_{ij}^{c}$ and $\bm M_{ij}^{h}$ to optimize the student model to make the extracted features more clear. It can be described by:
\begin{equation}
\begin{aligned}
    \bm L_{ISL}(\bm F_{S})=\frac{1}{B^{2}}\|\bm M^{c}-\bm M^{h}\|_{1}
\end{aligned},
\end{equation}
 From the perspective of the student model, the clear intrinsic similarity matrix contains much more pedestrian information. ISL can bridge the gap caused by hazy weather and make the student model have a more robust feature representation.

 \begin{figure}[htbp]
	\centering
	\includegraphics[width=2.7in,height=2.0in]{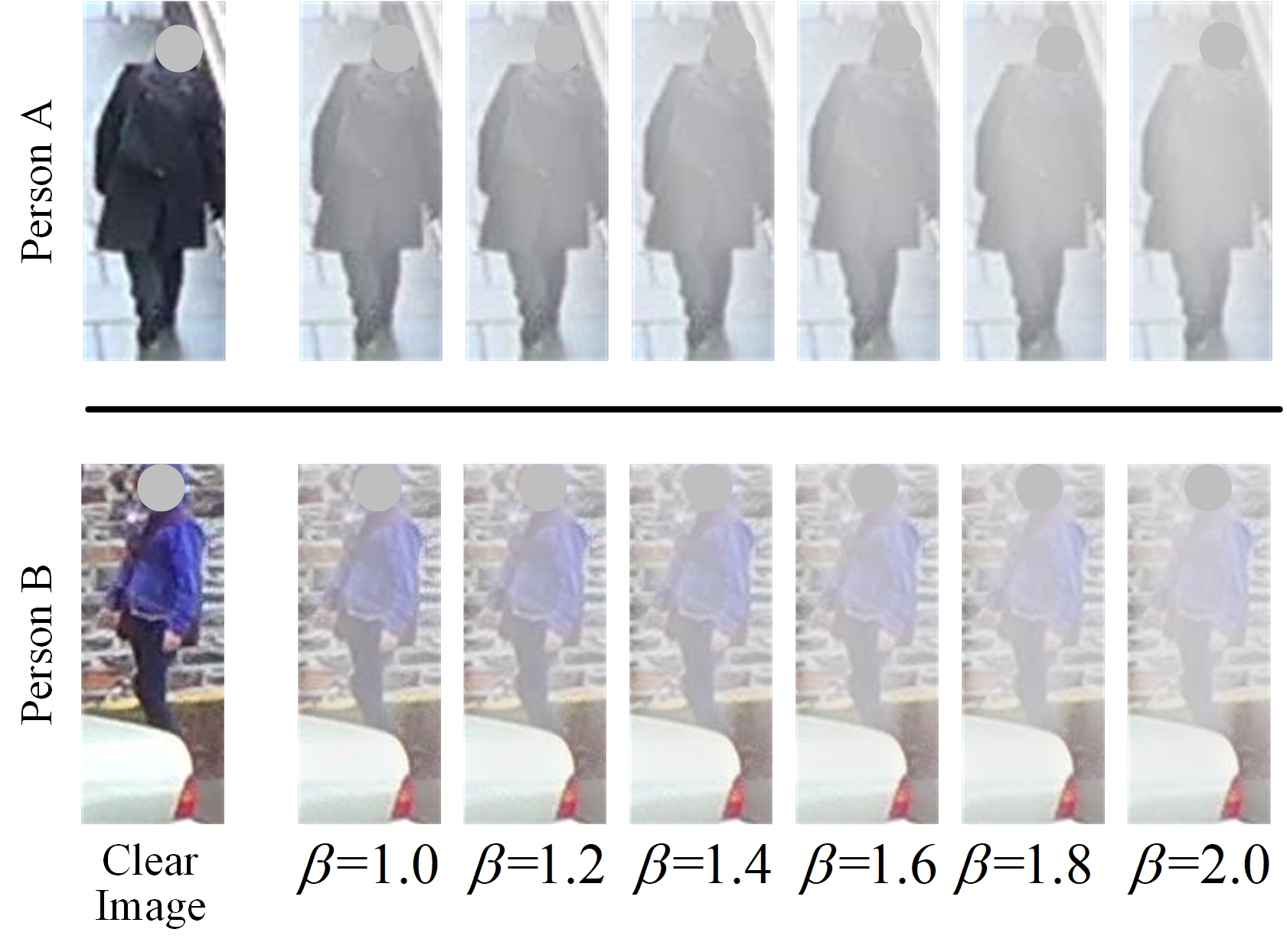}
	\caption{comparison among generated hazy images from different values of $\beta$. It can be seen that the hazy density in the image is basically close to the real scene.}
	\label{fig:4}
\end{figure}

\subsection{Interference Distillation by Adversarial Learning}
 ISL uses the divergence produced by interference at the feature level to guide the student model to learn robust feature representations. However, it does not satisfy the model interfered by hazy. The distribution of hazy images contains noises, which lead to a domain gap between the distribution of hazy and clear images.
 The interference can be distilled by forcing the hazy distribution closer to the clear distribution. To this end, the adversarial training is adopted between the student model and the discriminator. The discriminator is fed with clear feature $\bm f_{i}^{T}$ and hazy feature $\bm f_{i}^{S}$ to obtain the clear clues $\bm h_{i}^{T}=\bm D(\bm f_{i}^{T})$ and interference clues $\bm h_{i}^{S}=\bm D(\bm f_{i}^{S})$ by optimizing the cross-entropy loss:
\begin{equation}
\begin{aligned}
    \bm L_{D\_CE}(\bm D)=-\bm ylog \hat{\bm y}
\end{aligned},
\end{equation}
 where $\bm y$ represents the one-hot label (whether contain hazy interference), $\hat{\bm y}$ is the softmax output of the classifier corresponding to the input image. It can be implied that the discriminator can capture the hazy interference clues from hazy image features when the discriminator trained with the classification loss can successfully distinguish whether the feature contains interference. In order to eliminate the noise in the hazy data distribution, KL-divergence loss is used to optimize the student model for interference distillation under the guidance of the clear feature clues, which can be expressed as:
\begin{equation}
\begin{aligned}
    \bm L_{IDKL}(\bm F_{S})=\delta^{2}\bm KL (softmax(\bm h^{T}/\delta)\|softmax(\bm h^{S}/\delta))
\end{aligned},
\end{equation}
 where $\bm h^{T}$ and $\bm h^{S}$ represent the interference and clear information, respectively.

 The final loss for student network is the combination of following items, i.e., cross entropy loss $\bm L_{CE}$ in Equation~(1), intrinsic similarity loss $\bm L_{ISL}$ in Equation~(4), interference distillation loss $\bm L_{IDKL}$ in Equation~(6):
\begin{equation}
\begin{aligned}
    \bm L=\bm L_{CE} + \lambda_{1}\bm L_{ISL} + \lambda_{2}\bm L_{IDKL}
\end{aligned},
\end{equation}
where $\lambda_{1}$ and $\lambda_{2}$ are hyper-parameters used to control the importance of the two latter losses.

\begin{table*}[!ht]\small
 	\centering {\caption{ Experimental results of the proposed methods compared with the state-of-the-art methods on the hazy dataset.}
 		\renewcommand\arraystretch{1.4}
 			\begin{tabular}{|c|c|c|c|c|c|c|c|c|c|c|}
 				\hline
 				\multirow{2}*{~Methods~} & \multicolumn{4}{c|}{Duke $\rightarrow$ Hazy-Market} & \multicolumn{4}{c|}{Market1501 $\rightarrow$ Hazy-Duke} \\
 				\cline{2-9}
 				&~Rank-1~ &~Rank-5~&~Rank-10~&~~~mAP~~~&~Rank-1~ &~Rank-5~&~Rank-10~&~~~mAP~~~\\
 				\hline
 				ECN(CVPR'19)~\cite{Zhong2019} & 28.6&47.8&57.2&10.0&30.7&45.4&52.2&12.3\\
 				\hline
 				PDA-Net(ICCV'19)~\cite{Li2019} & 27.9& 47.1& 56.3& 9.7& 29.1&43.9&51.1&11.5\\
 				\hline
 				UDA(PR'20)~\cite{Song2020b} &27.3&44.8&53.7&11.0&21.5&34.5&40.0&8.1\\
 				\hline
 				PAST-ReID(ICCV'19)~\cite{Zhang2019} & 29.5& 47.9& 55.3&11.9& 26.5&39.5&45.1&10.8\\
 				\hline
 				SSG(ICCV'19)~\cite{Fu2019} & 29.1& 48.1& 55.8& 12.4& 25.8&37.2&42.5&10.6\\
 				\hline
 				ACT(AAAI'20)~\cite{Yang2020} &28.7&46.3&55.0&11.5&24.2&36.1&42.3&9.9\\
 				\hline
 				MMT-500(ICLR'20)~\cite{Ge2020} &29.9&48.1&56.0&12.4&26.9&40.3&46.0&11.7\\
 				\hline
 				\textbf{Ours}  &\bf48.4&\bf66.4&\bf72.9&\bf21.9&\bf37.7&\bf53.9&\bf59.7&\bf20.8\\
 				\hline
 		\end{tabular}
 	}
 	\label{tab:1}
 \end{table*}

\section{Experiments}

\subsection{Synthetic Hazy Dataset and Evaluation}

 The hazy image is formed through a widely used model proposed by Koschmieder \textit{et al}. ~\cite{Koschmieder1924}:
\begin{equation}
\begin{aligned}
    I(x)=J(x)t(x)+A(1-t(x))
\end{aligned},
\end{equation}
where $I(x)$ and $J(x)$ denote the generated hazy image and the clear image, respectively. $A$ represents the atmospheric light, which describes the intensity of scattered light in the scene at each channel of an image. $A$ is set to 0.9 in this paper empirically. The scene transmission $t(x) $ express attenuation in intensity due to scattering.
\begin{equation}
\begin{aligned}
    t(x)=e^{-\beta d(x)}
\end{aligned},
\end{equation}
 where $\beta$ is the medium extinction coefficient whose value represents the hazy density. $d(x)$ is the depth map corresponding to the clear image $J(x)$.

 The monocular depth estimation method proposed by Godard \textit{et al}.~\cite{Godard2019} is employed in this work to generate depth map. The hazy images are finally generated by combining the generated depth map with the original clear image through Equation~(8) and Equation~(9). The visualization is shown in Fig.\ref{fig:3}.

 In order to make the hazy density~\cite{Zhang2020} of the generated images in the test set closer to the real scene, we set the value range of $\beta$ sampled from an uniform distribution $\mathcal{U}(1,2)$ whose lower and upper boundary are set to 1 and 2 respectively. The generated hazy images with different values of $\beta$ are shown in Fig.\ref{fig:4}. It can be seen that when $\beta=1$, the density of hazy is shallow, and discriminative clues such as the appearance and bags of the person A can be clearly seen. With the increase of $\beta$, the denser the hazy makes the entire pedestrian disappear gradually so that the appearance and other clues become inconspicuous. To sum up, the value of $\beta$ is set between 1 and 2 to include the different hazy density in the real scene, which is in line with the practice application.

 The method is therefore tested on the following two synthetic hazy datasets: 1) The Hazy-Market1501 includes 32,668 generated hazy images of 1,501 pedestrians captured by six, where 12,936 labeled images of 751 pedestrians for training, and 19,732 images of 750 pedestrians for testing; 2) The Hazy-DukeMTMC-reID contains 16,522 images of 702 pedestrians for training, 2,228 images as query and 17,661 images as gallery. For simplicity, Hazy-Market1501 and Hazy-DukeMTMC-reID are referred as Hazy-Market and Hazy-Duke respectively in the rest of the paper. The Cumulative Match Characteristic (CMC) and the mean Average Precision (mAP) are used as the evaluation protocols.

 \begin{table*}[!ht]\small
 	\centering{\caption{ Ablation study on Market$\rightarrow$Hazy-Duke and Duke$\rightarrow$Hazy-Market (\%). \textit{Baseline} trained on labeled source dataset with $\bm L_{ID}$. $\bm L_{ISL}$ loss function used in intrinsic similarity learning. $\bm L_{KL}$ loss function used in interference distillation.}
 		\renewcommand\arraystretch{1.3}
 			\begin{tabular}{|c|c|c|c|c|c|c|c|c|c|c|}
 				\hline
 				\multirow{2}*{Methods} & \multirow{2}*{Source} & \multicolumn{4}{c|}{Target: Hazy-Duke} &\multirow{2}*{Source} & \multicolumn{4}{c|}{Target: Hazy-Market}\\
 				\cline{3-6}
 				\cline{8-11}
 				& &~~R-1~~&~~R-5~~&~R-10~&~mAP~& &~~R-1~~&~~R-5~~&~R-10~&~mAP~\\
 				\hline
 				\textit{Baseline} ($\bm L_{ID}$) &\multirow{4}*{Market} &15.2&22.3&37.1&5.8&\multirow{4}*{Duke}&22.6&35.1&40.1&7.5\\
 				\textit{Baseline}+$\bm L_{ISL}$&&36.5&53.1&59.3&20.6&&47.7&65.9&72.7&21.8\\
 				\textit{Baseline}+$\bm L_{IDKL}$&&26.3&40.3&46.3&12.1&&34.4&54.2&61.5&13.6 \\
 				\textit{Baseline}+$\bm L_{ISL}$+$\bm L_{IDKL}$&&\bf37.7&\bf53.9&\bf59.7&\bf20.8&&\bf48.4&\bf66.4&\bf72.9&\bf21.9 \\
 				\hline
 		\end{tabular}
 	}
 	\label{tab:3}
 \end{table*}

\subsection{Implementation Details}
 \textbf{Pre-training in source domain.} The same architecture is used for both the teacher and student models: a ResNet-50~\cite{He2016} pre-trained on ImageNet as the backbone. The source domain is trained in a supervised way for 120 epochs with the batch-size of 64. The initial learning rate is set to 0.00035 and decreased to 1/10 of its previous value at the 40th and 90th epoch in total 120 epochs. \textbf{Adaptation in target domain.} The clear images of the target domain are used to generate the hazy images to help the model better adapt to the hazy environment. The student model is trained through Equation~(4) and Equation~(6) for 60 epochs, while the discriminator is optimized via Equation~(5). $\epsilon$, $\delta$ are set to 0.1, 10. For the final loss function, $\lambda_{1}$ and $\lambda_{2}$ are set to 4 and 0.1, respectively.

 \begin{figure}[!ht]
 	\centering
 	\includegraphics[width=3.0in,height=2.3in]{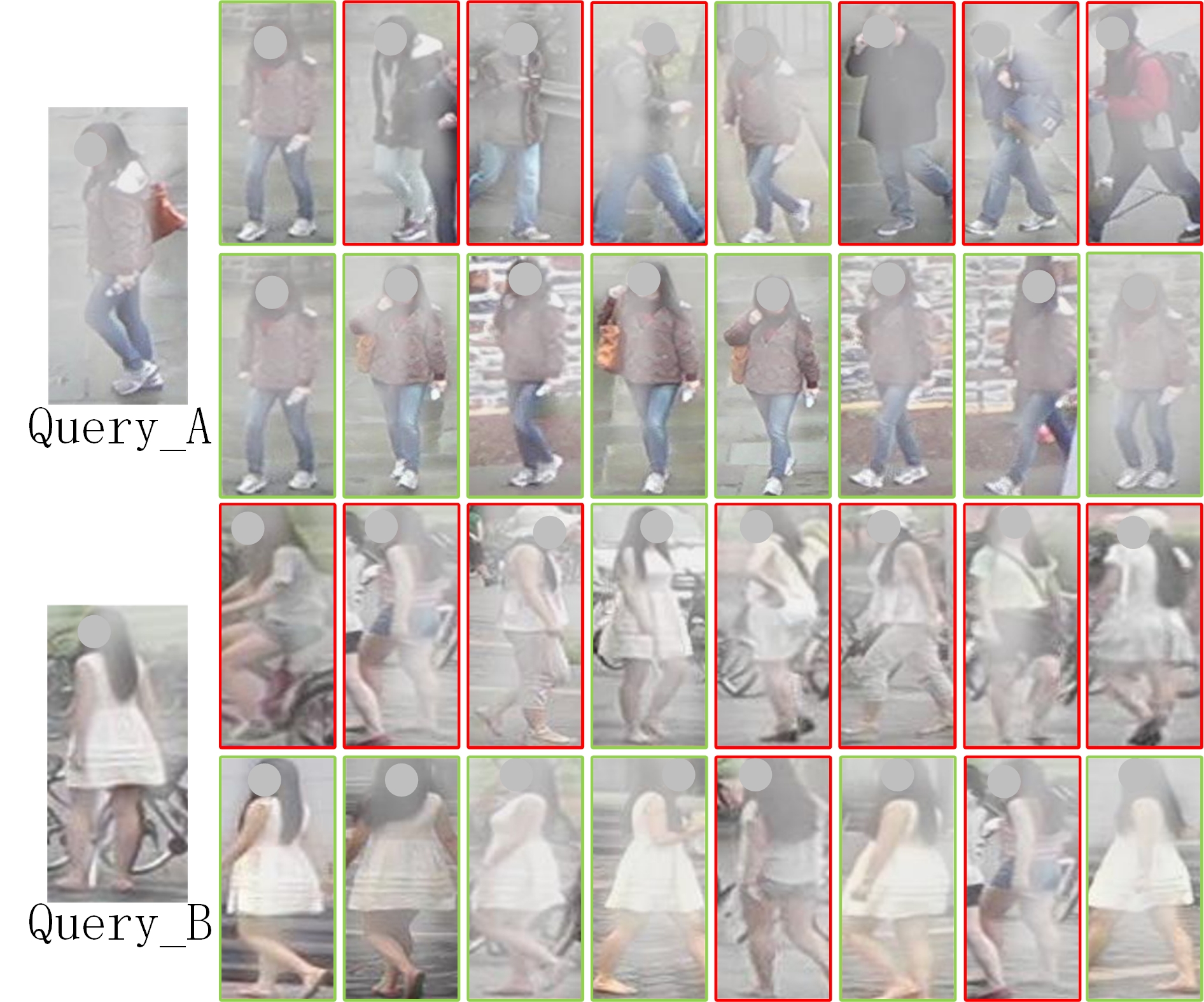}
 	\caption{Visualization of two ranking lists. The query images A and B are showed with the top-8 ranking samples on the right. In each case, the first row is obtained by \textit{Baseline}, and the second row by \textit{ISM}. Green box indicates that the example of the same identity with the query where the red box indicates the one of different identity.}
 	\label{fig:5}
 \end{figure}

\subsection{Comparison with the state-of-the-art methods}

 In the comparison with other methods, the training images of all compared methods are clear regardless of whether they are from the source domain or the target domain. In test, the compared models are directly evaluated on the hazy test dataset of target domain. In Table 1, the proposed ISM is compared with the state-of-the-art domain adaptation methods (ECN~\cite{Zhong2019}, PDA-Net~\cite{Li2019}) and clustering-based methods (UDA~\cite{Song2020b}, PAST-ReID~\cite{Zhang2019}, SSG~\cite{Fu2019}, ACT~\cite{Yang2020}, MMT-500~\cite{Ge2020}). All experimental results are based on the source code released by those methods.

 One can observe that the Rank-1 and mAP achieved by ISM are superior to those methods. In particular, the proposed model achieves 48.4\%/21.9\% for Duke~\cite{Ristani2016}$\rightarrow$Hazy-Market, 37.7\%/20.8\% for Market~\cite{Zheng2015}$\rightarrow$Hazy-Duke in Rank-1/mAP. The state-of-the-art domain adaption method ECN~\cite{Zhong2019} is newly released and the proposed ISM surpass it by a large margin on Rank-1 and mAP in all tasks. Compared with the best clustering-based method MMT-500~\cite{Ge2020} that achieves 85.3\%/68.5\% (72.9\%/57.6\%) for the task of Duke$\rightarrow$Market (Market$\rightarrow$Duke) in Rank-1/mAP, the proposed model outperforms it by 18.5\%/9.5\%(10.8\%/9.1\%) when the Hazy-Market (Hazy-Duke) is used as the test dataset.

 \begin{figure}[!ht]
 	\centering
 	\includegraphics[width=3.2in,height=2.4in]{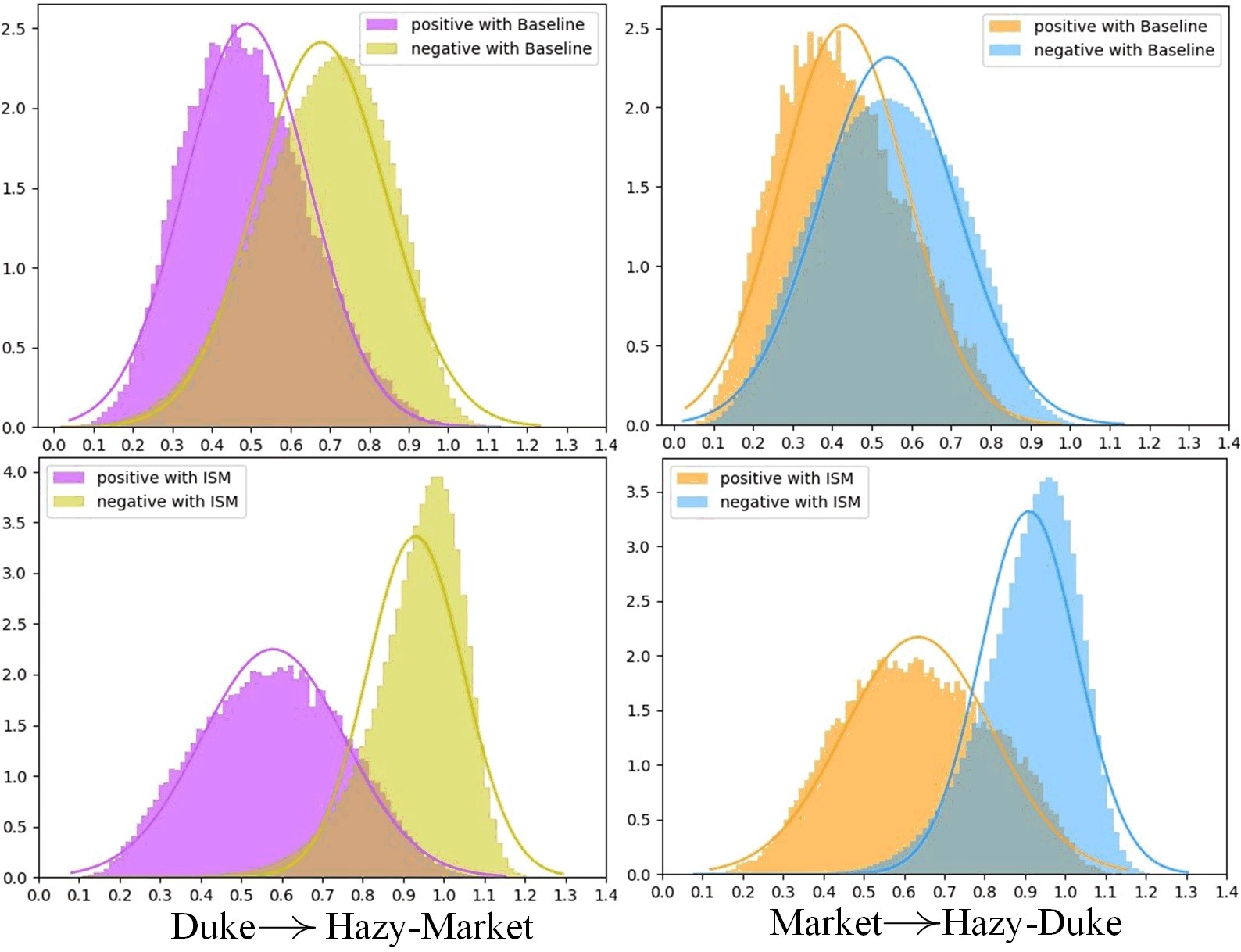}
 	\caption{The similarity scores distribution of positive and negative samples. The first row and the second row are obtained by \textit{Baseline} and \textit{ISM}, respectively. The more overlap of the two distributions, the poorer discriminative of the model.}
 	\label{fig:6}
 \end{figure}

\subsection{Ablation Study}
 As mentioned previously, the proposed approach consists of two parts:
 intrinsic similarity learning ($\bm L_{ISL}$) and interference distillation by KL-divergence ($\bm L_{IDKL}$). In this section ,we evaluate the effectiveness of each part on two tasks of Duke$\rightarrow$Hazy-Market and Market$\rightarrow$Hazy-Duke. The experimental results are compared in Table 2. The \textit{Baseline} denote only using source domain with $\bm L_{ID}$ to train the model. The results show that the proposed $\bm L_{ISL}$ and $\bm L_{IDKL}$ are effective for reducing the hazy interference on both Duke$\rightarrow$Hazy-Market and Market$\rightarrow$Hazy-Duke. It achieves the highest performance level by complementing each other.

 \begin{figure}[!t]
 	\centering
 	\includegraphics[width=3.2in,height=1.8in]{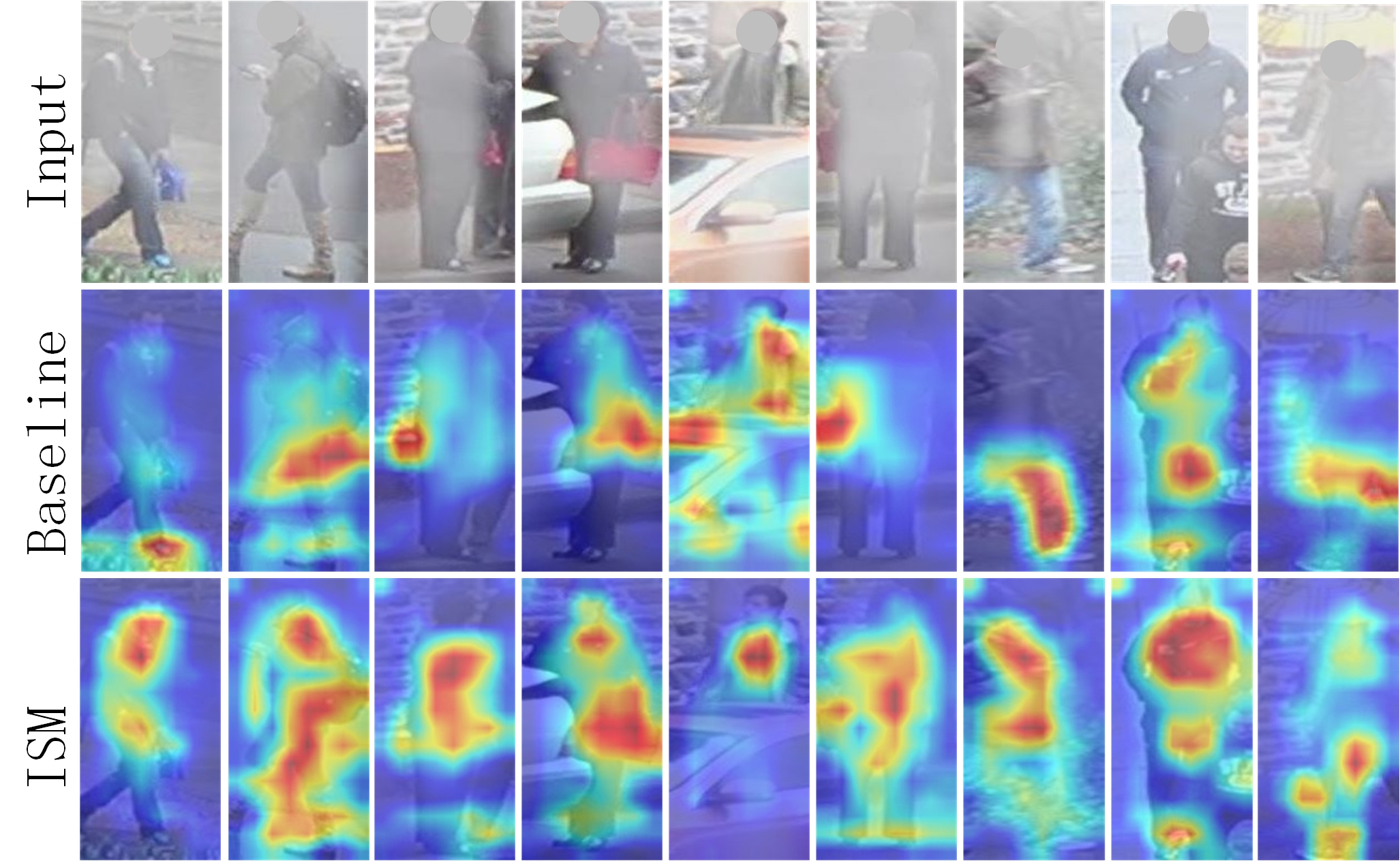}
 	\caption{Visualization of attention maps. From comparison, the \textit{ISM} pays more attention to the whole part of pedestrians.}
 	\label{fig:7}
 \end{figure}

\subsection{Further Analysis}

 \paragraph*{Qualitative Results of Ranking List.}
  As can be seen from Fig.\ref{fig:5}, the \textit{Baseline} model fails to retrieve most of correct results under hazy condition. On the contrary, ISM achieves competitive retrieval performance even for those pedestrian images with significant hazy interference.

 \paragraph*{Analysis of Similarity Distribution.}
 In Fig.\ref{fig:6}, the distribution is obtained by calculating the mean and variance of the similarity scores of positive and negative samples. The insufficient discrimination of the model between positive and negative samples will cause the two data distributions to be close. In the tasks Duke$\rightarrow$Hazy-Market and Market$\rightarrow$Hazy-Duke, the overlap of the two distributions obtained by \textit{ISM} is much smaller than that of the \textit{Baseline}, which implies the \textit{ISM} can effectively distinguish the positive and negative samples.

 \paragraph*{Visualization of Attention Map.}
 To further investigate the differences between the teacher model and the student model, we support the above experimental results by adopting Grad-Cam~\cite{Selvaraju2017} to visualize their outputs. In Fig.\ref{fig:7}, the attention map of baseline model is not ideal. For examples, the results in columns 1, 3, and 9 show that the attention of the network is not concentrated in the discriminative part of the pedestrians due to the hazy interference that disturbs the feature representation. After domain adaptation, it can be seen that, in the third row, the student model can adapt to the hazy weather and pay more attention to the whole part of pedestrians and details. The above visualizations prove that the proposed ISM can extract robust features that are not contaminated by hazy.

\section{Conclusions}
In this work, we consider the interference of hazy on the unsupervised person Re-ID, and contribute two synthetic hazy datasets for further study. We also proposed a novel model to reduce the interference caused by hazy weather for the person Re-ID task. The experimental results on two synthetic hazy datasets showed that the proposed model is superior to the compared state-of-the-art domain adaptation models. The ablation study verified the effectiveness of each sub-module. Further analysis proved the proposed model is more discriminative than the benchmark, and it gives more attention to the details of pedestrian. In future study, the general model could be investigated for a variety of inclement whether conditions.

\bibliographystyle{IEEEbib}
\bibliography{pj-icme2021}

\end{document}